\documentclass[sigconf]{acmart}
\usepackage{graphicx}
\usepackage{subcaption}
\usepackage{multirow}
\usepackage{makecell}
\usepackage[table]{xcolor}
\usepackage[normalem]{ulem}
\usepackage{algorithm}
\usepackage{algpseudocode}
\usepackage{array}

\usepackage{savesym}
\usepackage{amsmath}
\savesymbol{Bbbk} 
\usepackage{amssymb} 

\usepackage{xcolor}

\AtBeginDocument{%
  }

\copyrightyear{2026}
\acmYear{2026}
\setcopyright{cc}
\setcctype{by}
\acmConference[MM '26]{Proceedings of the 34th ACM International Conference on Multimedia}{November 10--14, 2026}{Rio de Janeiro, Brazil}
\acmBooktitle{Proceedings of the 34th ACM International Conference on Multimedia (MM '26), November 10--14, 2026, Rio de Janeiro, Brazil}
\acmDOI{10.1145/3767308.3835311}
\acmISBN{979-8-4007-2213-4/2026/11}

\acmSubmissionID{2326}

\begin{document}


\title{HiRS-Agent: A Hierarchical Multi-Agent System for Reliable Long-Horizon Remote Sensing Task Solving}


\author{Boyang Mu}
\orcid{0009-0008-1854-714X}
\affiliation{%
\institution{Beijing University of Posts and Telecommunications}
  \department{State Key Laboratory of Networking and Switching Technology}
  \city{Beijing}
  \country{China}
}
\email{muboyang@bupt.edu.cn}

\author{Zhiwei Wei}
\orcid{0000-0002-3494-3686}
\affiliation{%
   \institution{Hunan Normal University}
  \department{School of Geographic Sciences}
  \city{Changsha}
  \state{Hunan}
  \country{China}
}
\email{2011301130108@whu.edu.cn}

\author{Mugen Peng}
\orcid{0000-0002-4755-7231}
\affiliation{%
    \institution{Beijing University of Posts and Telecommunications}
  \department{State Key Laboratory of Networking and Switching Technology}
  \city{Beijing}
  \country{China}
}
\email{hy\_wang@bupt.edu.cn}

\author{Wenjia Xu}
\orcid{0000-0002-1425-4162}
\authornote{Corresponding author.}
\affiliation{%
    \institution{Beijing University of Posts and Telecommunications}
  \department{State Key Laboratory of Networking and Switching Technology}
  \city{Beijing}
  \country{China}
}
\email{xuwenjia@bupt.edu.cn}



\begin{abstract}
	Recent advances in large language models and multimodal models have pushed remote sensing (RS) processing from simple perception models to agentic systems designed to tackle complex, long-horizon RS tasks. However, existing systems often rely on monolithic decision-making frameworks, which fail to accommodate the multi-stage, interdependent nature of RS tasks. This centralized approach leads to challenges such as unstable task execution, incorrect tool usage, and error propagation across stages.
	To address these issues, we propose \textbf{HiRS-Agent}, a hierarchical multi-agent system for long-horizon RS task solving. HiRS-Agent adopts a two-level collaborative architecture: the Manager Layer handles dynamic routing, step-level verification, replanning, and termination control, while the Specialist Layer organizes domain-specific tools according to the RS workflow and is responsible for subtask reasoning and tool execution. To further enhance the system's capability, we introduce a two-stage supervised tuning strategy and a verification-guided hierarchical reinforcement learning stage to jointly optimize coordination and tool-use policies. Experiments on Earth-Agent Benchmark and ThinkGeo show that HiRS-Agent substantially improves long-horizon tool-use capability and final-task correctness, demonstrating the effectiveness of structured multi-agent collaboration for reliable RS agents. The code is publicly available at
\url{https://github.com/IntelliSensing/HiRS-Agent}.
\end{abstract}



\begin{teaserfigure}
  \centering
  \includegraphics[width=0.95\textwidth]{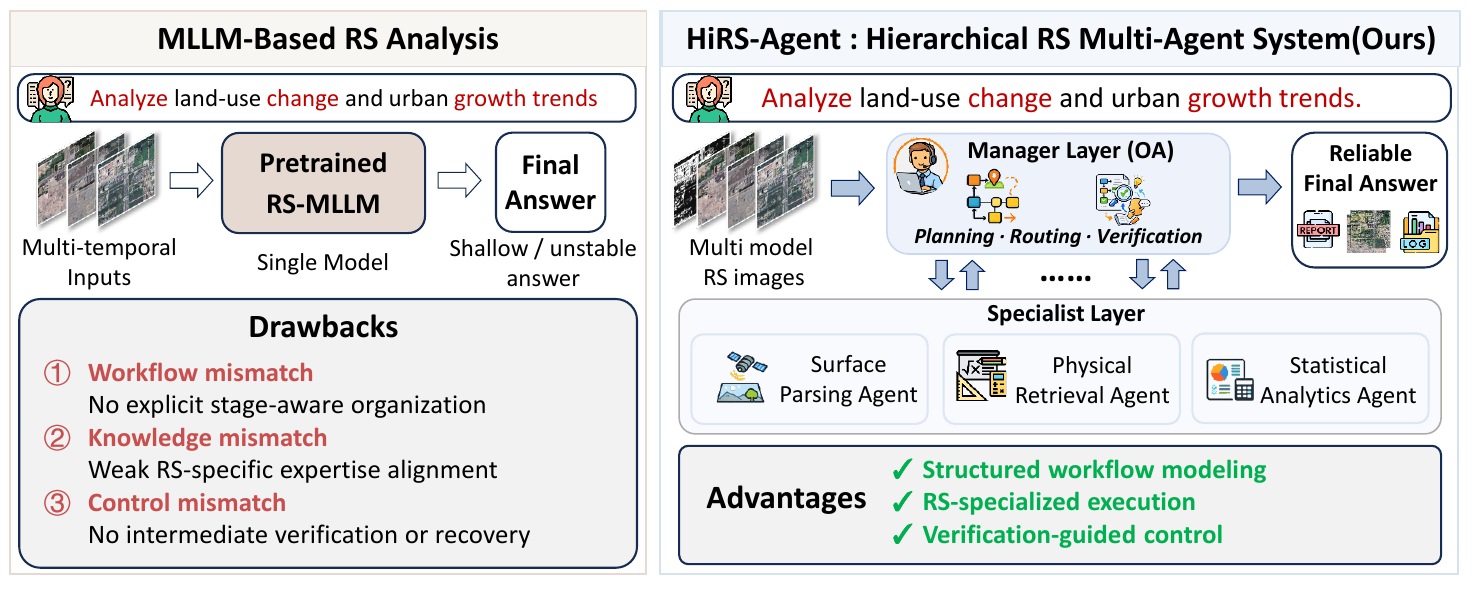}
  \Description{Comparison between generic MLLM-based RS agents and our HiRS-Agent for long-horizon remote sensing tasks. Given the same complex query, generic models suffer from workflow, knowledge, and control mismatches, leading to shallow and unstable outputs. In contrast, HiRS-Agent organizes multi-temporal RS inputs through a Manager–Specialist hierarchical architecture with verification-guided control, enabling structured workflow execution and reliable long-horizon reasoning.}
  \caption{\textbf{Overview of HiRS-Agent.} Generic MLLM-based RS agents (left) suffer from workflow, knowledge, and control mismatches in long-horizon tasks, while HiRS-Agent (right) enables structured and reliable execution via a Manager–Specialist hierarchy with verification-guided control.}
  \label{fig:teaser}
\end{teaserfigure}


\maketitle

\section{Introduction}
\label{sec:intro}

Remote sensing (RS) has become a fundamental infrastructure for earth observation, environmental monitoring, disaster assessment, and geospatial intelligence. With the rapid growth of observation platforms and sensor modalities, modern RS applications increasingly rely on large-scale, heterogeneous, and multimodal data, making analysis workflows more complex and labor-intensive~\cite{DL-in-RS,RS-servey}. In parallel, multimodal large language models (MLLMs) have shown strong capabilities in perception-oriented RS tasks, such as visual question answering~\cite{GeoChat,LHRS-Bot}, scene understanding~\cite{RemoteCLIP,SkyScript,SkyEyeGPT,BTCChat}, object detection~\cite{EarthGPT}, and semantic segmentation~\cite{RSVLM,RS-ChatGPT}. However, many real-world RS applications go far beyond one-shot perception. They often require multi-step reasoning, tool invocation, intermediate-result interpretation, and iterative decision making across a long processing chain. This trend is pushing RS systems from static perception models toward agentic frameworks capable of autonomously organizing and executing RS workflows.

Recent efforts in this direction can be roughly viewed from two complementary perspectives. The first line focuses on RS-oriented multimodal assistants and foundation models, which strengthen RS-specific perception, language grounding, and instruction following~\cite{LHRS-Bot,RS-ChatGPT}. The second line moves further toward executable RS agents and tool-augmented systems, exploring structured tool use, multi-step planning, geospatial reasoning, and workflow-oriented evaluation~\cite{RS-Agent,Earth-Agent}. Nevertheless, existing progress mainly demonstrates the feasibility of agentizing RS tasks, while reliable long-horizon execution in realistic RS workflows remains insufficiently addressed. The main difficulty is that many existing RS agents are mainly adapted from general-purpose agent paradigms rather than designed around the intrinsic structure of RS workflows~\cite{RS-Agent}. In practice, RS tasks often consist of multiple interdependent stages, forming long processing chains with strong upstream--downstream coupling. For instance, in a flood-mapping task, an error in the early parsing stage, such as mistaking cloud shadows for water, can directly affect downstream water-index computation, inundation-area estimation, and final reporting. This stage-dependent nature makes RS task solving especially sensitive to the quality of intermediate results, which must remain consistent with physical, spectral, and spatial constraints throughout the workflow.

Against this background, existing RS-agent systems suffer from three major mismatches. \textbf{Workflow mismatch}: although RS tasks follow structured processing chains, many current systems still rely on generic agent designs that do not explicitly model stage dependency or long-horizon workflow organization. \textbf{Knowledge mismatch}: the LLM serving as the ``brain'' of the agent is usually not sufficiently aligned with RS expertise, and therefore does not truly understand RS-specific physical mechanisms, spectral constraints, or professional data-processing procedures. \textbf{Control mismatch}: existing frameworks mainly emphasize local action selection or final-answer evaluation, while providing limited support for intermediate-state verification and recovery from execution failures. These limitations make reliable long-horizon RS task solving fundamentally challenging.

To address these three mismatches, we propose \textbf{HiRS-Agent}, a Hierarchical Remote Sensing Multi-Agent System for reliable long-horizon RS task solving. Unlike generic agent templates, HiRS-Agent is designed around the structured and stage-dependent nature of RS workflows. It adopts a two-level architecture with a \textbf{Manager Layer} for global planning and workflow control, and a \textbf{Specialist Layer} for domain-specialized RS reasoning and tool execution. To improve reliability, the Manager Layer further verifies intermediate results under RS-specific constraints and can reroute, replan, or repair the workflow when necessary. And the Specialist Layer is structured according to the RS processing chain (spectral parsing $\rightarrow$ physical retrieval $\rightarrow$ spatial analytics), with tools grouped into stage-aligned subsets. In this way, HiRS-Agent provides workflow-aware organization, RS-specialized execution, and verification-guided control for long-horizon RS tasks.

Beyond the architecture, we further optimize HiRS-Agent with a dedicated training pipeline. To address the knowledge mismatch between natural-language instructions and professional RS procedures, we construct an RS-agent instruction corpus and introduce \textbf{Expert-to-Workflow Alignment Tuning} (\textbf{Expert-tuning}), a two-stage supervised tuning strategy that first injects RS expertise and then aligns high-level task intent with executable RS workflows. Building on this architecture, we further develop \textbf{Verification-Guided Hierarchical Reinforcement Learning} (\textbf{VG-HRL}) to jointly optimize global orchestration and local tool execution under sparse long-horizon supervision. Together, these designs improve the stability and reliability of RS-agent execution.

In summary, our main contributions are as follows:
\begin{itemize}
    \item \textbf{A hierarchical multi-agent framework for long horizon RS task solving.}
    We propose HiRS-Agent, a two-level multi-agent system tailored to the structured and stage-dependent nature of RS workflows.

    \item \textbf{Verification-guided workflow control.}
    We introduce a memory-aware step-level verification mechanism that supports rerouting, replanning, and recovery under RS-specific constraints during multi-stage execution.

    \item \textbf{Workflow-aware training for RS agents.}
    We develop Expert-tuning for RS expertise injection and workflow alignment, and further propose VG-HRL to jointly optimize global coordination and local tool execution.

    \item \textbf{Systematic evaluation on representative benchmarks.}
    Experiments on Earth-Agent Benchmark (Earth-Bench)~\cite{Earth-Agent} and ThinkGeo~\cite{ThinkGeo} show that HiRS-Agent improves long-horizon RS task execution and final-task correctness, especially on lightweight open-source backbones.
\end{itemize}

\begin{figure*}[t]  
  \centering        
  \includegraphics[width=0.95\textwidth]{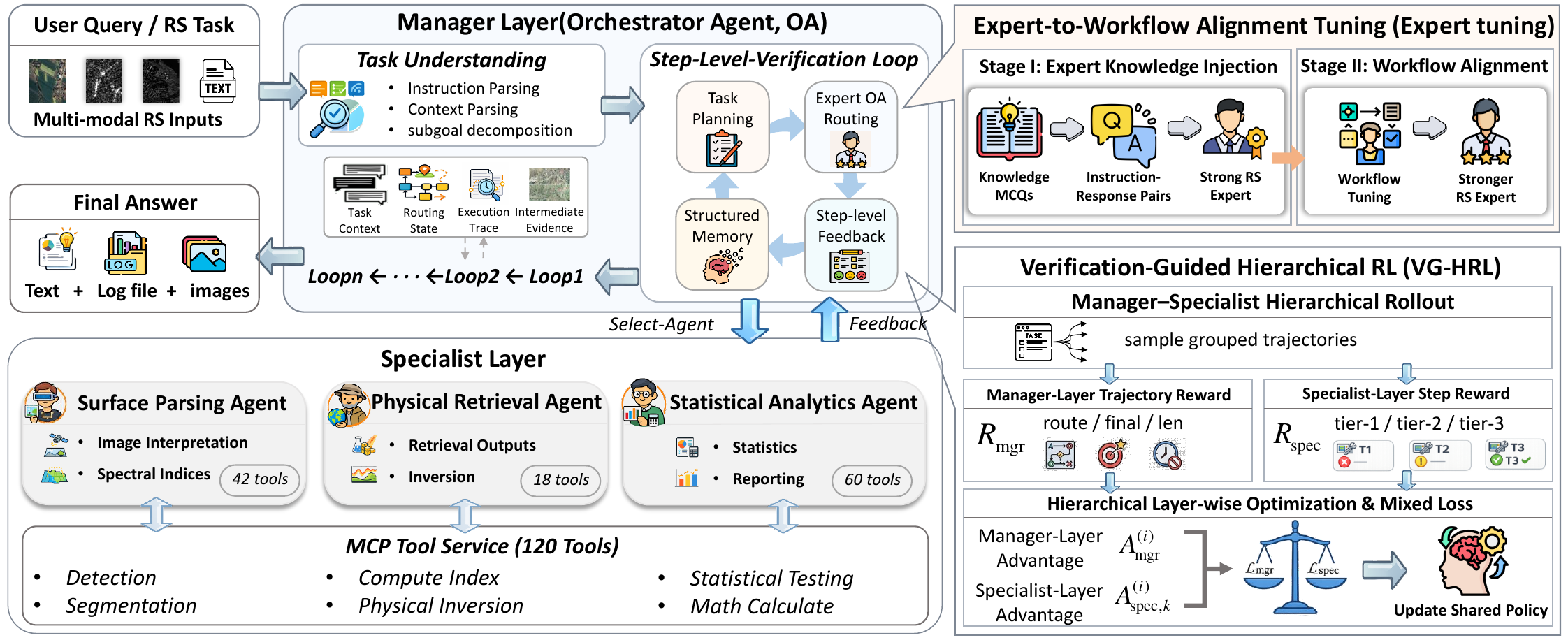}
  \caption{Overview of HiRS-Agent: A Hierarchical Multi-Agent System for Long-Horizon RS Task Solving.}
  \Description{A Hierarchical Framework for Tool-Augmented Remote Sensing Task Solving}
  \label{fig:overview} 
\end{figure*}

\section{Related Work}

\subsection{LLM Agents and RS Agents}

Recent LLM agents have extended language models from one-shot generation to iterative reasoning, planning, tool use, and feedback refinement. ReAct~\cite{ReAct} demonstrated the effectiveness of interleaving reasoning and acting, while Reflexion~\cite{Reflexion} improved agent behavior through feedback-conditioned verbal reflection. Related methods such as plan-first reasoning~\cite{Plan-and-Solve} and Tree-of-Thoughts~\cite{Tree-of-Thoughts} further show that explicit process modeling can improve long-horizon decision quality beyond direct prompting.

Building on these advancements, RS-oriented MLLMs have significantly enhanced perception and instruction following in Earth observation. Representative systems such as RS ChatGPT~\cite{RS-ChatGPT}, LHRS-Bot~\cite{LHRS-Bot}, SkyEyeGPT~\cite{SkyEyeGPT}, EarthGPT~\cite{EarthGPT}, and RingMo-Agent~\cite{RingMo-Agent} explored RS-specific multimodal alignment, instruction tuning, and unified reasoning across heterogeneous sensors and platforms~\cite{huang2026structural,huang2025investigating,huang2026dissecting}.  More recent work extended this line toward executable RS agents and
tool-augmented workflows, exploring multi-step tool
use~\cite{RS-Agent}, geospatial reasoning~\cite{GeoAgent},
interactive change interpretation~\cite{Change-Agent}, workflow
management~\cite{CangLing-KnowFlow}, tool
creation~\cite{zhao2026}, and reasoning trace alignment~\cite{OpenEarthAgent}, while Earth-Agent~\cite{Earth-Agent} and ThinkGeo~\cite{ThinkGeo} provided benchmark evaluation for structured tool use and multi-step planning in RS.

Despite this progress, reliable long-horizon execution remains insufficiently addressed. Existing RS systems have made advances in perception, workflow coverage, and tool integration, yet many still rely on shallow execution chains, static prompts, fixed templates, or limited recovery mechanisms. Even when hierarchical decomposition is introduced, step-level verification and state-adaptive rerouting are rarely treated as first-class control signals~\cite{XIJIAO-MAS}. As a result, early errors can still propagate across the workflow and compromise downstream analytical validity, leaving a critical gap for long-horizon RS task solving.

\subsection{Multi-Agent and RL Optimization}
Multi-agent architectures have been increasingly adopted to improve modularity, role specialization, and coordination in complex reasoning and task-solving systems. Prior work on debate-style collaboration~\cite{Debate} showed that multiple agents can improve reasoning diversity and factuality, while geospatial studies such as GeoAgent~\cite{GeoAgent} demonstrated the usefulness of hierarchical coordinator--worker organizations for spatial analysis. HTAM~\cite{XIJIAO-MAS} further argued that agents in specialized domains should be organized according to intrinsic task dependency graphs rather than loosely mimicking human social roles. For RS tasks, the key issue is therefore not simply whether multiple agents are used, but whether the system can maintain execution coherence under evolving intermediate states~\cite{GeoColab,CangLing-KnowFlow}. These observations suggest that long-horizon RS workflows require hierarchical and state-adaptive coordination mechanisms that can route subtasks to specialized experts and maintain consistency across stages.

Recent RL studies on tool-using LLMs suggest that final-answer supervision is too coarse for complex agent behavior~\cite{Agent-as-Tool}. ToolRL~\cite{ToolRL}, ReTool~\cite{ReTool}, AgentPO~\cite{Multi-Agent-Collaboration-via-Reinforcement-Learning}, and related work~\cite{MURKA} showed that process-level optimization can improve strategic tool invocation, collaboration behavior, and hierarchical decision policies. This need is especially strong in RS, where effective long-horizon execution depends not only on final correctness, but also on routing quality, tool validity, parameter accuracy, execution efficiency, and recovery behavior across the full trajectory. However, few existing studies combine hierarchical coordination, process-level verification, and fine-grained RL optimization within a unified framework for reliable long-horizon RS task execution. HiRS-Agent is designed to address this gap through a workflow-aware hierarchical architecture and verification-guided optimization.

\section{Methodology}

HiRS-Agent is designed to address the three mismatches identified in Sec.~\ref{sec:intro}, namely workflow mismatch, knowledge mismatch, and control mismatch. To this end, our methodology comprises three components. First, we introduce a \emph{hierarchical agent architecture} that organizes RS task solving according to the dependency structure of RS workflows. Second, we develop \emph{Expert-to-Workflow Alignment Tuning} to align the foundation model with RS expertise and executable processing procedures. Third, we propose \emph{Verification-Guided Hierarchical Reinforcement Learning} to optimize global orchestration and local tool execution under long-horizon supervision.

RS tasks generally unfold as structured processing workflows in which early-stage decisions directly affect downstream outcomes. Following the canonical RS progression from \emph{spectral parsing} to \emph{physical retrieval} and then to \emph{statistical analytics}, we design HiRS-Agent as a two-level collaborative system composed of a \textbf{Manager Layer} and a \textbf{Specialist Layer}, as illustrated in Fig.~\ref{fig:overview}. The Manager Layer maintains global workflow coherence through planning, routing, and verification, while the Specialist Layer performs domain-specialized reasoning and tool execution at different workflow stages.

A typical execution of HiRS-Agent proceeds as follows. Given a user instruction and multimodal RS inputs, the Manager Layer formulates sub-goals and constraints, routes each sub-goal to an appropriate specialist agent, and verifies the returned intermediate result under RS-specific constraints. The workflow then proceeds through iterative execution, verification, and adaptive correction until task completion. The following subsections detail these components.

\subsection{Hierarchical Agent Architecture}
\label{sec:orche}

\subsubsection{Manager Layer: Memory-Aware Control and Verification}
\label{OA}
The Manager Layer is instantiated as a single \textbf{Orchestrator Agent (OA)}, which serves as the core controller of HiRS-Agent. 
Beyond conventional task planning, the OA is designed to support task decomposition, history-aware verification, and adaptive execution control, which are essential for handling inter-stage dependencies and preventing error propagation. This design is motivated by the nature of RS tasks, where early-stage decisions (e.g., spectral parsing or band selection) can directly affect downstream physical inversion and statistical analysis.

\textbf{Global Memory for Long-Horizon Dependency Modeling.}
To explicitly model long-horizon dependencies in RS workflows, the OA maintains a structured global state
\begin{equation}
	s_t = \{x, d, g_t, \mathcal{H}_t, C_t\},
\end{equation}
where $x$ denotes the instruction, $d$ represents multimodal RS input data, $g_t$ is the current sub-goal, $\mathcal{H}_t$ is a structured global memory, and $C_t$ represents dynamic constraints. Unlike conventional execution logs, $\mathcal{H}_t$ is organized as a structured and queryable memory that stores: (i) task context, (ii) routing states, (iii) execution traces, and (iv) intermediate evidence.

\textbf{Task Planning and History-Aware Verification under RS Constraints.}
To support both task planning and history-aware verification under RS constraints, the OA outputs routing and control decisions
\begin{equation}
	(r_t, c_t) \leftarrow \pi_{\theta}(s_t; \rho_{\mathrm{OA}}),
\end{equation}
where $\rho_{\mathrm{OA}}$ denotes the OA role prompt, $r_t$ selects the appropriate specialist agent in the Specialist Layer, and $c_t$ determines the next action (e.g., continue, replan, verify, or terminate). Since HiRS-Agent adopts a shared-parameter architecture, the OA and specialist agents are prompt-conditioned role instantiations of the same backbone policy $\pi_\theta$, distinguished only by role-specific prompts. In addition to sub-goal decomposition and routing, the OA performs history-aware verification, which constitutes its core functionality.

After each specialist-agent execution, the OA receives structured step-level feedback, including tool calls, outputs, and execution evidence. Instead of validating results solely based on local outputs, the OA retrieves relevant global memory
\begin{equation}
m_t = \mathrm{Retrieve}(\mathcal{H}_t, g_t, C_t),
\end{equation}
and evaluates the current step by jointly considering the intermediate result, prior execution traces, and domain constraints.

To enforce RS-specific constraints during intermediate execution, the verification process is defined as
\begin{equation}
z_t = V(g_t,o_t,C_t,m_t), \qquad  
z_t \in \{\text{pass},\text{uncertain},\text{fail}\},
\end{equation}
where $o_t$ is the current step output. The verifier performs three complementary checks:
(i) schema validity, ensuring required fields and types are satisfied,
(ii) sub-goal consistency, ensuring the output advances or fulfills the current objective, and
(iii) constraint satisfaction, ensuring that the result remains physically, spectrally, and statistically plausible under $C_t$ and prior memory $m_t$.
In RS workflows, this includes verifying physical consistency, such as valid inversion relationships and reasonable value ranges, spectral correctness, such as proper band usage and index computation, and statistical plausibility, such as reasonable aggregation patterns and output distributions.

\textbf{Verification-Guided Adaptive Control.}
Based on the verification outcome, the OA performs adaptive control to maintain workflow consistency and correct execution errors. Specifically, \textbf{FAIL} triggers replan-on-failure, including tool repair, alternative routing, or sub-goal adjustment; \textbf{UNCERTAIN} triggers cross-checking through re-execution or alternative reasoning paths; and \textbf{PASS} commits the result into memory and proceeds to the next step.

Importantly, all intermediate results, including failures and recovery traces, are stored in $\mathcal{H}_t$, enabling subsequent decisions to be history-aware rather than myopic. This closed-loop design of global memory, history-aware verification, and adaptive control directly addresses the long-horizon dependency and constraint-driven nature of RS tasks, allowing the OA to dynamically adjust execution strategies, explicitly model inter-stage dependencies, and effectively mitigate error accumulation.

\begin{figure*}[t]
    \centering
    \begin{subfigure}[b]{0.307\textwidth}
        \centering
        \includegraphics[width=\linewidth]{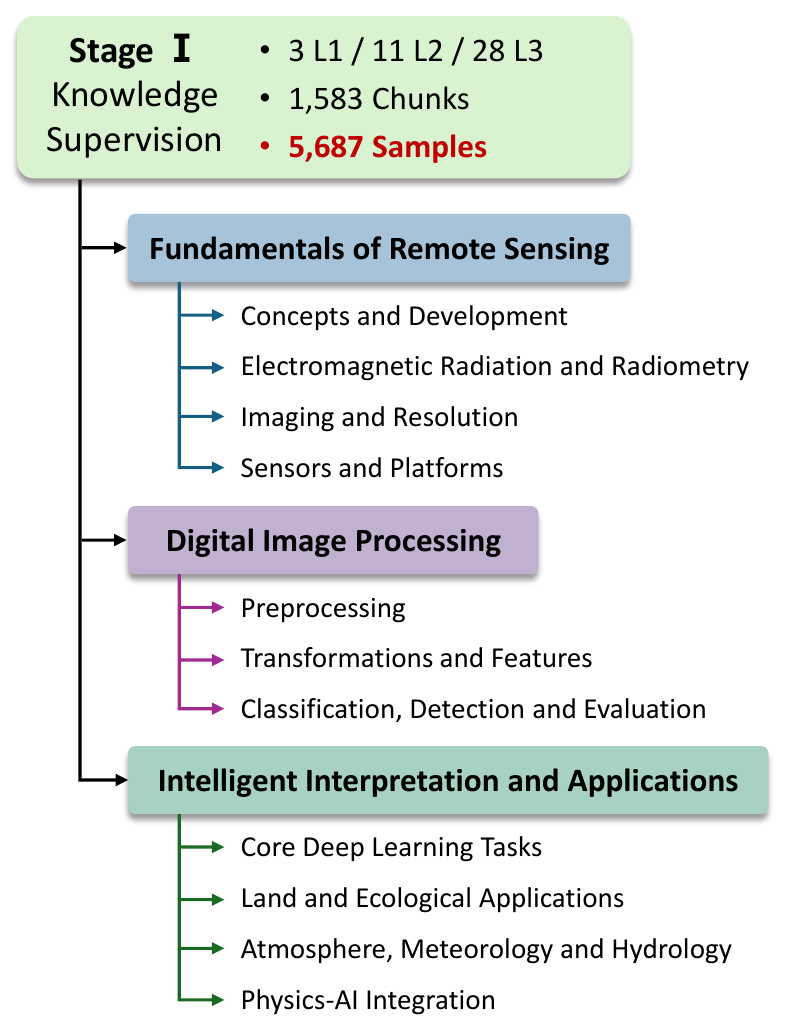}
        \caption{Knowledge Supervision}
        \label{fig:Knowledge}
    \end{subfigure}
    \hfill
    \begin{subfigure}[b]{0.285\textwidth}
        \centering
        \includegraphics[width=\linewidth]{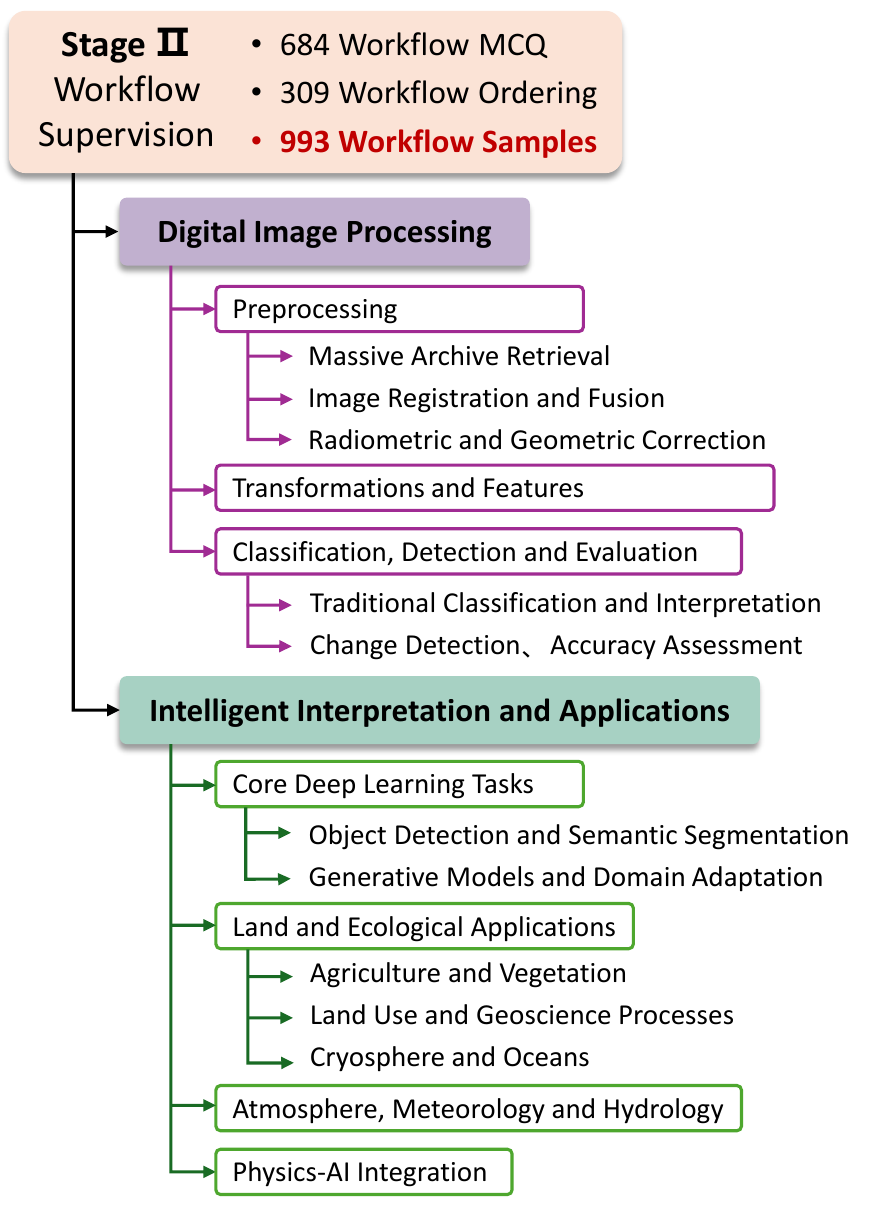}
        \caption{Workflow Supervision}
        \label{fig:Workflow}
    \end{subfigure}
    \hfill
    \begin{subfigure}[b]{0.320\textwidth}
        \centering
        \includegraphics[width=\linewidth]{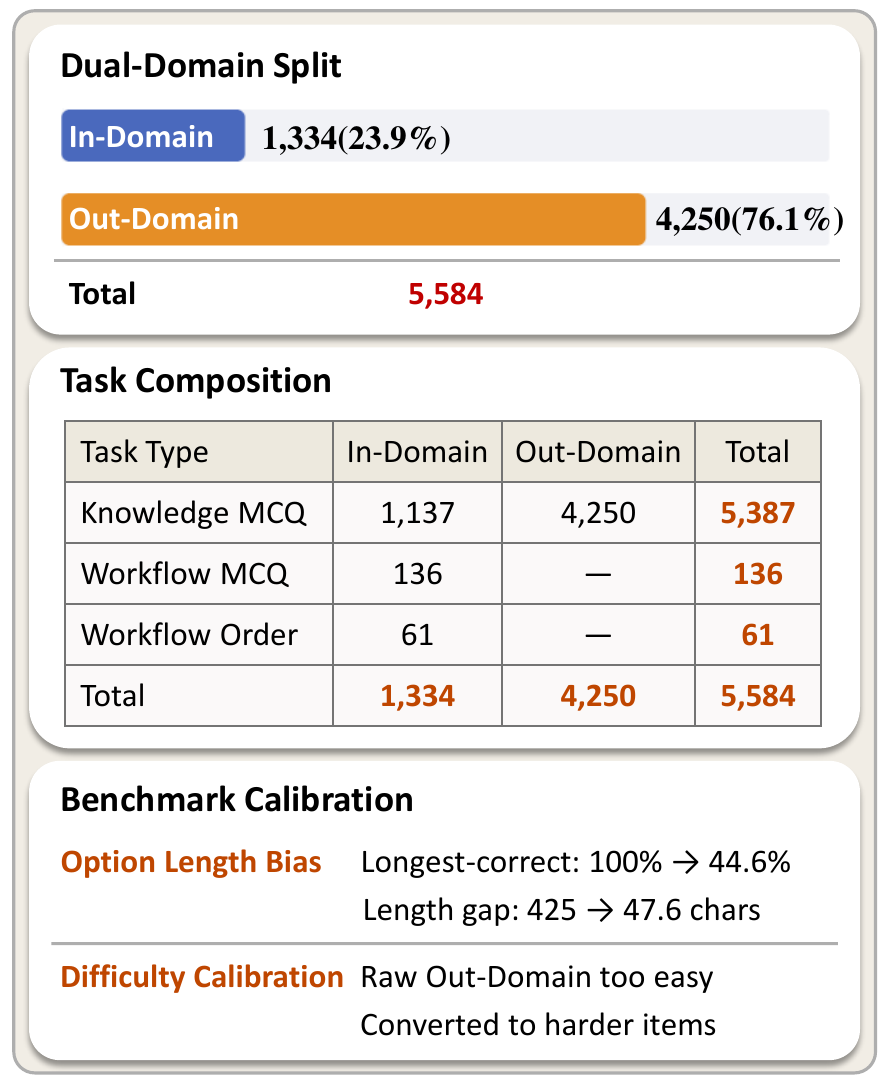}
        \caption{RS-EXPERT-BENCHMARK}
        \label{fig:Benchmark}
    \end{subfigure}
    \caption{Taxonomy-aligned composition of the two-stage Expert-tuning supervision data and the statistics of RS-EXPERT-BENCHMARK.}
    \label{fig:Expert_tuning_supervision_data}
    \Description{Taxonomy-aligned composition of the two-stage Expert tuning supervision data and the statistics of RS-EXPERT-BENCHMARK.}
\end{figure*}

\subsubsection{Specialist Layer: Domain-Specialized Planning and Tool Use}

The Specialist Layer consists of domain-specialized agents that transform OA-issued sub-goals into tool-level actions. Given $(g_t, \mathcal{H}_t, C_t)$, each specialist agent outputs a structured tool call $u_t=(\texttt{tool}, \texttt{args})$, whose execution returns an output $o_t$ for verification and an optional step summary for $\mathcal{H}_t$. The OA dynamically routes among specialist families and applies step-level verification after each execution.

To support stable long-horizon execution and effective routing, the Specialist Layer is organized into three function-oriented expert families: \textbf{Surface Parsing Agent} (SPA), \textbf{Physical Retrieval Agent} (PRA), and \textbf{Statistical Analytics Agent} (SAA). This design follows the functional progression of RS workflows from spectral parsing and physical inversion to statistical analysis. By partitioning the tool space, each specialist operates over a focused and domain-consistent set of tools, reducing cross-domain confusion and improving execution reliability. Incorrect tool invocation, including wrong tool selection and malformed arguments, further motivates the step-level verification and repair loop described in Sec.~\ref{OA}.

\textbf{SPA} handles perception- and parsing-oriented sub-goals, including object detection, segmentation, ROI or mask generation, and spectral-index-oriented parsing (e.g., NDVI/NDWI cues). Its tool space focuses on vision parsing, spatial region processing, and index computation primitives, making it suitable for early-stage scene understanding and region-level RS image analysis. SPA returns $u_t=(\texttt{tool}, \texttt{args})$ with optional summaries such as region statistics or mask quality indicators. Typical failure modes include cloud or shadow interference, boundary ambiguity, and spatial misalignment across modalities.

\textbf{PRA} handles physical retrieval and inversion sub-goals, such as LST-oriented tasks and quantitative RS product generation. Its tool space covers physically grounded retrieval operators, calibration and normalization routines, and lightweight quality-control checks. PRA returns $u_t=(\texttt{tool}, \texttt{args})$ and optional diagnostics (e.g., confidence or quality flags) for downstream verification. Typical failure modes include unit or range inconsistency, sensitivity to atmospheric artifacts, and unstable estimates under low-quality observations.

\textbf{SAA} handles data-centric sub-goals such as data aggregation and cleaning, statistical testing and correlation analysis, and report-level metric computation and comparison. Its tool set focuses on tabular and time-series processing, statistical inference, and controlled file I/O for intermediate artifacts, thereby supporting higher-level result summarization and decision making. SAA returns $u_t=(\texttt{tool}, \texttt{args})$ and may provide structured summaries (e.g., tables, test results, or comparative indicators). Typical failure modes include schema mismatches across intermediate outputs, missing-value propagation, and violated test assumptions.

\subsection{Expert-to-Workflow Alignment Tuning}
\label{sec:Expert tuning}

To align general large language models with workflow-driven RS tasks, we propose \textbf{Expert-tuning}, a two-stage supervised fine-tuning strategy. \textbf{Stage I} injects structured RS domain knowledge, while \textbf{Stage II} aligns natural language instructions with executable RS processing procedures.

\textbf{Stage I: Basic Knowledge Alignment.}
We first construct a structured RS knowledge system for agent-oriented reasoning. The knowledge space follows a three-level taxonomy with 3 L1 topics, 11 L2 modules, and 28 L3 domains, covering \emph{Fundamentals of RS}, \emph{Digital Image Processing}, and \emph{Intelligent Interpretation and Applications}. We collect and curate knowledge from four authoritative RS textbooks and map the extracted content onto the predefined L1--L3 structure. Based on this taxonomy, we build the RS-EXPERT corpus through five stages: content extraction, normalization, chunking, rewriting, and annotation, resulting in 1,583 semantically coherent knowledge chunks. From this corpus, we further construct 5,687 knowledge-centric training samples, as shown in Fig.~\ref{fig:Knowledge}.

\textbf{Stage II: Workflow Alignment.}
Stage II constructs workflow supervision from real RS application pipelines collected from multi-agent demo logs and verified by two experts. It focuses on two workflow-relevant taxonomy branches, \emph{Digital Image Processing} and \emph{Intelligent Interpretation and Applications}, which involve multi-step tool invocation and procedural reasoning. As illustrated in Fig.~\ref{fig:Workflow}, the data comprise two formats: \emph{Workflow MCQ} for selecting correct processing plans and \emph{Workflow Order} for recovering tool-use sequences. Stage II contains 684 Workflow MCQ and 309 Workflow Order samples (993 in total), spanning 14 task families. Each workflow is tagged with its taxonomy branch to connect the Stage I knowledge system with executable procedures.

For supervised fine-tuning, 80\% of the data are converted into instruction-response pairs, with the remaining 20\% reserved for evaluation. To assess knowledge acquisition and workflow awareness beyond format memorization, we further construct \emph{RS-EXPERT-BENCHMARK}. It evaluates knowledge grounding and workflow generalization using held-out in-domain data and additional out-of-domain samples from unseen \emph{RS Handbook} sources. The dual-domain split, task composition, and benchmark calibration are summarized in Fig.~\ref{fig:Benchmark}.

\subsection{Verification-Guided Hierarchical RL}

\label{sec:hrl_optimization}

To further improve workflow-level decision quality beyond imitation learning, we introduce \textbf{VG-HRL}, a verification-guided post-SFT reinforcement learning strategy for HiRS-Agent. The key idea is to jointly optimize the Manager and Specialist Layer according to their distinct roles in long-horizon RS workflows. Specifically, we design (1) a \textbf{hierarchical reward scheme} that separates trajectory-level coordination from step-level tool execution, and (2) a \textbf{role-aware grouped optimization strategy} that aligns credit assignment and advantage estimation with the two-level collaboration. In implementation, we instantiate this optimization with Group Relative Policy Optimization (GRPO)~\cite{GRPO}.

\textbf{Hierarchical Reward Design.}
The Manager Layer is responsible for trajectory-level coordination, while the Specialist Layer performs step-level tool execution. To avoid signal confusion and credit-assignment mismatch, we define separate rewards for the two levels and combine them only at the loss level. Throughout this section, quantities marked with a hat denote normalized or scaled reward components, while unmarked terms denote raw feedback signals.

For the Manager Layer, we use a trajectory-level reward:
\begin{equation}
	R_{\mathrm{mgr}}
	=
	\omega_{\mathrm{route}}\hat R_{\mathrm{route}}
	+
	\omega_{\mathrm{final}}\hat R_{\mathrm{final}}
	+
	\omega_{\mathrm{len}}\hat R_{\mathrm{len}},
\end{equation}
where $\hat R_{\mathrm{route}}$ measures normalized routing correctness, $\hat R_{\mathrm{final}}$ evaluates normalized final task success, and $\hat R_{\mathrm{len}}$ penalizes unnecessarily long trajectories after scaling. Their explicit definitions are provided in the supplementary material.

For the Specialist Layer, we replace binary failure gating with a \emph{domain-aware graded gating reward}. Each specialist agent is associated with a domain tool registry, and each tool call is categorized into one of three tiers, for the $k$-th tool call in the $i$-th rollout:
\begin{equation}
	R_{\mathrm{spec},k}^{(i)}=
	\begin{cases}
		-1.0, & \tau_k = 0,\\
		-0.5 + 0.2R_{\mathrm{exec}}, & \tau_k = 1,\\
		0.5 + 0.2R_{\mathrm{exec}} + 0.3\hat R_{\mathrm{args}}, & \tau_k = 2,
	\end{cases}
	\label{eq:spec_reward}
\end{equation}
Here, $\tau_k=0$, $1$, and $2$ correspond to Tier 1 (hallucinated out-of-domain tool), Tier 2 (wrong but in-domain tool), and Tier 3 (correct tool), respectively. In Eq.~\eqref{eq:spec_reward}, $R_{\mathrm{exec}}\in\{-1,0,+1\}$ denotes raw execution feedback, and $\hat R_{\mathrm{args}}\in[0,1]$ denotes the normalized argument-quality score, computed only when the predicted tool is correct. By distinguishing hallucinated out-of-domain tools from semantically closer in-domain mistakes, this reward preserves more informative within-group variance for stable relative optimization and improves local credit assignment. Intermediate normalization details are deferred to the supplementary material.

\begin{algorithm}[t]
\caption{VG-HRL Optimization for HiRS-Agent}
\label{alg:vghrl_optimization}
\small
\begin{algorithmic}[1]
\Require Shared policy $\pi_\theta$, reference policy $\pi_{\mathrm{ref}}$, task set $\mathcal{D}$, group size $G$, weight $\alpha=0.5$
\For{each minibatch $\mathcal{B}\subset\mathcal{D}$}
    \State $\mathcal{L}_{\mathrm{mgr}},\mathcal{L}_{\mathrm{spec}}\gets 0$
    \For{each task $x\in\mathcal{B}$}
        \State Sample $G$ complete trajectories $\{y^{(i)}\}_{i=1}^{G}$ using HiRS-Agent
        \State Compute Manager-Layer rewards $\{R_{\mathrm{mgr}}^{(i)}\}_{i=1}^{G}$
        \State Compute Specialist-Layer step rewards $\{R_{\mathrm{spec},k}^{(i)}\}$
        \State Compute Manager-Layer advantages $\{A_{\mathrm{mgr}}^{(i)}\}_{i=1}^{G}$ by grouping rollouts of the same task
        \State Compute Specialist-Layer advantages $\{A_{\mathrm{spec},k}^{(i)}\}$ by grouping rewards with the same $(x,k)$
        \State Accumulate manager loss $\mathcal{L}_{\mathrm{mgr}}$
        \State Accumulate specialist loss $\mathcal{L}_{\mathrm{spec}}$
    \EndFor
    \State $\mathcal{L}_{\mathrm{mixed}} \gets \lambda \mathcal{L}_{\mathrm{mgr}} + (1-\lambda)\mathcal{L}_{\mathrm{spec}}$
    \State Update $\theta$ by minimizing $\mathcal{L}_{\mathrm{mixed}}$
\EndFor
\State \Return optimized shared policy $\pi_\theta$
\end{algorithmic}
\end{algorithm}

\begin{table*}[t]
	\centering
	\small
	\caption{Main results on Earth-Bench. \textbf{Bold} / \underline{underline}
denote best / second-best. Unless noted, all reproduced rows share the same
Qwen3-4B checkpoint (Expert-tuned + VG-HRL), tool registry, and evaluation setup.}
	\label{tab:main_table}
	\setlength{\tabcolsep}{2.9pt}
	\renewcommand{\arraystretch}{1.04}
	\begin{tabular}{l*{12}{c}}
		\toprule
		\multirow{2}{*}{Model} &
		\multicolumn{2}{c}{Tool-Any-Order$\uparrow$} &
		\multicolumn{2}{c}{Tool-In-Order$\uparrow$} &
		\multicolumn{2}{c}{Tool-Exact-Match$\uparrow$} &
		\multicolumn{2}{c}{Param-Match$\uparrow$} &
		\multicolumn{2}{c}{Efficiency$\downarrow$} &
		\multicolumn{2}{c}{Accuracy$\uparrow$} \\
		\cmidrule(lr){2-3}\cmidrule(lr){4-5}
		\cmidrule(lr){6-7}\cmidrule(lr){8-9}
		\cmidrule(lr){10-11}\cmidrule(lr){12-13}
		& AP & IF & AP & IF & AP & IF &
		AP & IF & AP & IF & AP & IF \\
		\midrule

		\multicolumn{13}{l}{\itshape Closed-source models} \\

		GPT-5 &
		\underline{69.11} & \underline{71.16} &
		\underline{58.25} & \underline{60.68} &
		45.57 & 45.79 &
		26.10 & 25.80 &
		2.4146 & 2.8428 &
		\textbf{65.59} & \textbf{63.16} \\

		Gemini-2.5 &
		57.96 & 61.80 &
		45.44 & 50.78 &
		31.86 & 40.92 &
		17.49 & 23.54 &
		3.0246 & 2.4709 &
		\underline{54.66} & \underline{55.06} \\

		GPT-4o &
		65.73 & 66.88 &
		50.70 & 53.20 &
		\underline{46.17} & \underline{47.47} &
		\underline{26.71} & \underline{27.77} &
		\underline{2.1579} & 2.6088 &
		45.34 & 44.94 \\

		\midrule
		\multicolumn{13}{l}{\itshape Open-source models} \\

		DeepSeek-V3.1 (37B) &
		\textbf{78.40} & \textbf{77.98} &
		\textbf{63.25} & \textbf{64.33} &
		\textbf{48.89} & \textbf{50.01} &
		\textbf{30.76} & \textbf{31.34} &
		2.6283 & 2.6624 &
		50.20 & 52.23 \\

		Qwen3-32B &
		39.76 & 42.39 &
		21.56 & 33.79 &
		9.51 & 26.10 &
		8.13 & 17.73 &
		2.7274 & \underline{1.9010} &
		20.62 & 24.80 \\

		Qwen3-8B &
		36.19 & 44.95 &
		16.73 & 31.05 &
		5.26 & 19.10 &
		2.96 & 10.49 &
		5.5726 & 5.0779 &
		18.55 & 14.92 \\

		Qwen3-4B &
		20.47 & 29.69 &
		1.32 & 10.89 &
		0.00 & 8.63 &
		0.00 & 4.21 &
		3.9747 & 5.4554 &
		15.73 & 10.08 \\

		\midrule
		\multicolumn{13}{l}{
			\itshape Generic agent architectures
			(shared Qwen3-4B checkpoint)
		} \\

		CoT~\cite{Chain-of-thought} &
		25.97 & 27.90 &
		20.96 & 27.97 &
		5.83 & 8.78 &
		16.76 & 22.58 &
		7.0321 & 6.7015 &
		4.84 & 6.05 \\

        \makecell[l]{ReAct~\cite{ReAct}} &
		27.45 & 35.82 &
		16.65 & 18.92 &
		6.84 & 19.42 &
		4.12 & 11.20 &
		4.8214 & 4.9125 &
		16.53 & 20.16 \\

		Plan\&Execute~\cite{Plan-and-Solve} &
		57.76 & 57.88 &
		36.24 & 36.18 &
		24.29 & 24.58 &
		10.82 & 10.90 &
		\textbf{1.3011} & \textbf{1.1771} &
		30.01 & 30.65 \\

		Debate~\cite{Debate} &
		47.50 & 47.53 &
		34.26 & 35.79 &
		1.53 & 1.81 &
		16.33 & 10.95 &
		13.3459 & 11.9874 &
		23.08 & 22.67 \\

		\midrule
		\multicolumn{13}{l}{
			\itshape RS-specific agent architectures
			(shared Qwen3-4B checkpoint)
		} \\

		RS-Agent~\cite{RS-Agent} &
		22.68 & 27.25 &
		15.84 & 18.61 &
		15.17 & 17.74 &
		8.90 & 9.18 &
		2.3385 & 2.2295 &
		29.44 & 33.06 \\

		OpenEarthAgent~\cite{OpenEarthAgent} &
		46.55 & 50.21 &
		39.01 & 41.68 &
		28.59 & 30.75 &
		18.98 & 20.75 &
		2.7099 & 2.7778 &
		18.15 & 19.76 \\

		\midrule
		\multicolumn{13}{l}{\itshape Proposed hierarchical multi-role system} \\

		\rowcolor{gray!15}
		HiRS-Agent (Qwen3-8B) &
		58.08 & 63.69 &
		45.94 & 53.10 &
		37.50 & 43.98 &
		20.82 & 26.83 &
		2.5248 & 3.0848 &
		48.39 & 53.62 \\

		\rowcolor{gray!15}
		HiRS-Agent (Qwen3-4B) &
		50.95 & 53.79 &
		41.10 & 46.34 &
		31.67 & 34.64 &
		19.54 & 20.81 &
		2.5059 & 3.3013 &
		43.95 & 45.56 \\

		\bottomrule
	\end{tabular}
\end{table*}

\textbf{Role-Aware Grouped Optimization.}
To match the hierarchical structure of HiRS-Agent, we estimate relative advantages at two different granularities. For the Manager Layer, all rollouts of the same task form one group:
\begin{equation}
	A_{\mathrm{mgr}}^{(i)}
	=
	\frac{
		R_{\mathrm{mgr}}^{(i)}-
		\mathrm{mean}(\{R_{\mathrm{mgr}}^{(j)}\}_{j=1}^{G})
	}{
		\mathrm{std}(\{R_{\mathrm{mgr}}^{(j)}\}_{j=1}^{G})+\epsilon
	}.
\end{equation}

For the Specialist Layer, rewards are grouped by (\emph{task, step index}), so that tool calls at the same execution position are compared against each other:
\begin{equation}
	A_{\mathrm{spec},k}^{(i)}
	=
	\frac{
		R_{\mathrm{spec},k}^{(i)}-
		\mathrm{mean}(\{R_{\mathrm{spec},k}^{(j)}\}_{j=1}^{G})
	}{
		\max\!\left(\mathrm{std}(\{R_{\mathrm{spec},k}^{(j)}\}_{j=1}^{G}),\epsilon_{\mathrm{th}}\right)
	}.
\end{equation}
Here, $\epsilon_{\mathrm{th}}$ is a fallback threshold for low-variance groups; further details are deferred to the supplementary material.

HiRS-Agent adopts a shared-parameter single-backbone architecture, where the Manager-Layer OA and all Specialist-Layer agents share the same LLM backbone and LoRA adapters and are distinguished only by role-specific prompts. Accordingly, $\mathcal{L}_{\mathrm{mgr}}$ and $\mathcal{L}_{\mathrm{spec}}$ denote the objectives computed from $A_{\mathrm{mgr}}^{(i)}$ and $A_{\mathrm{spec},k}^{(i)}$. Under this setting, the two levels are jointly optimized through
\begin{equation}
	\mathcal{L}_{\mathrm{mixed}}
	=
	\lambda \mathcal{L}_{\mathrm{mgr}}
	+
	(1-\lambda)\mathcal{L}_{\mathrm{spec}}.
\end{equation}

Importantly, the combination is performed at the loss level rather than the reward level, so that the two levels maintain independent normalization and stable optimization dynamics. In all HRL experiments, we set the mixed-loss weight to $\lambda=0.5$ as a balanced default, so that the Manager-Layer and Specialist-Layer losses contribute equally during joint optimization. Algorithm~\ref{alg:vghrl_optimization} summarizes the proposed VG-HRL optimization, where trajectory-level coordination and step-level tool execution are jointly optimized under a shared-parameter backbone.

\begin{table}[t]
	\centering
	\small
	\caption{Main results on ThinkGeo. \textbf{Bold} and \underline{underline} mark the best and second-best results. Step and E2E denote step-by-step and end-to-end metrics, respectively.} 
	\label{tab:main_table2}
	\begin{tabular}{l *{5}{c}} 
		\toprule
		\multirow{2}{*}{\textbf{Model}} &
		\multicolumn{3}{c}{Step Metrics$\uparrow$} &
		\multicolumn{2}{c}{E2E Metrics$\uparrow$} \\
		\cmidrule(lr){2-4} \cmidrule(lr){5-6}
		& Inst. & Tool. & Arg. & Ans. & Ans\_I \\
		\midrule
		\multicolumn{6}{l}{\itshape Closed-source models} \\
		GPT-4o & 73.31 & \underline{63.75} & \underline{33.31} & \underline{11.51} & \textbf{20.02} \\
		GPT-4-1106 & \textbf{82.44} & \textbf{73.21} & \textbf{37.74} & 9.46 & \underline{16.91} \\
		Claude-3.7-Sonnet & 21.35 & 26.21 & 0.33 & 8.95 & 11.42 \\
		
		\midrule
		\multicolumn{6}{l}{\itshape Open-source models} \\
		Qwen2.5-7b-Instruct & 57.38 & 45.63 & 18.84 & 6.91 & 9.28 \\
		Phi-3-mini-4k-Instruct & 18.40 & 16.88 & 9.65 & 7.16 & 6.12 \\
		Qwen3-8B & 20.98 & 13.36 & 3.26 & 7.67 & 8.68 \\
		Qwen3-4B & 18.35 & 8.54 & 1.24 & 6.07 & 7.79 \\
		
		\midrule
		\multicolumn{6}{l}{\itshape Proposed multi-agent system} \\
		\rowcolor{gray!15}
		HiRS-Agent (Qwen3-8B) & \underline{77.97} & 59.57 & 11.70 & \textbf{12.09} & 14.75 \\
		\rowcolor{gray!15}
		HiRS-Agent (Qwen3-4B) & 73.73 & 47.87 & 8.51 & 11.28 & 13.77 \\
		\bottomrule
	\end{tabular}
\end{table}

\begin{table}[t]
\centering
\caption{RS expertise gains and general capability retention after domain adaptation. \textbf{Bold} denote the best performance. }
\label{tab:rs_expertise_and_general}
\setlength{\tabcolsep}{3.8pt} 
\begin{tabular}{lccc}
\toprule
\textbf{Metric} & \makecell{\textbf{Qwen3-4B}\\\textbf{Base}} & \makecell{\textbf{Qwen3-4B}\\\textbf{Prompt}} & \makecell{\textbf{Qwen3-4B}\\\textbf{Expert-tuning}} \\
\midrule
\multicolumn{4}{l}{\itshape RS-EXPERT-BENCHMARK} \\
IN-Domain   & 76.43 & 80.13 & \textbf{95.84} \\
OUT-Domain  & 72.42 & 76.54 & \textbf{85.13} \\
Overall    & 73.35 & 77.37 & \textbf{87.60} \\
\midrule
\multicolumn{4}{l}{\itshape General Benchmarks} \\
MMLU-Redux~\cite{MMLU_Redux} & 66.57 & 66.53 & \textbf{66.80} \\
MATH-500~\cite{MATH-500}   & \textbf{67.80} & 67.60 & 66.20 \\
Multi-IF~\cite{Multi-IF}    & \textbf{58.35} & 58.30 & 55.53 \\
\bottomrule
\end{tabular}
\end{table}

\section{Experiments}

\subsection{Experimental Setup}

We evaluate HiRS-Agent using the official protocols of Earth-Bench and ThinkGeo \cite{Earth-Agent,ThinkGeo}, and additionally report domain adaptation on RS expertise and general-capability splits. Earth-Bench emphasizes long-horizon tool trajectories and end-to-end correctness, while ThinkGeo focuses on step-wise execution fidelity and final-answer quality. Following the benchmarks, we report two evaluation modes (AP and IF) and use the same tool interface, argument schema, and execution constraints across all methods. Baselines include both proprietary and open-source LLMs, and all methods are evaluated with the official scripts and metrics.

For training, HiRS-Agent uses a shared-parameter backbone, where all roles share the same Qwen3-4B or Qwen3-8B model and are differentiated only by role-specific system prompts and decoding settings. Both Expert-tuning and VG-HRL are implemented with LoRA and applied sequentially. Expert-tuning uses LoRA rank 8 and is trained for 3 epochs, while VG-HRL uses LoRA rank 64 with LoRA alpha 128 and is instantiated with GRPO using rollout group size $G=4$ and 3 training epochs per round. The mixed-loss weight is set to $\lambda=0.5$. Training is conducted on 2$\times$RTX 4090 24GB for Qwen3-4B and 4$\times$RTX 4090 24GB for Qwen3-8B. Additional qualitative results and detailed case studies are provided in the supplementary material.

\subsection{Main Results}
\label{sec:main_results}

Tables~\ref{tab:main_table} and~\ref{tab:main_table2} present the main results on Earth-Bench and ThinkGeo. Overall, HiRS-Agent consistently improves small open-source backbones on long-horizon RS agent tasks, showing that the gains mainly come from hierarchical specialization, verification-guided control, and post-SFT optimization rather than simply scaling up the backbone. As the backbone capacity increases from 4B to 8B, the proposed framework further narrows the gap to strong proprietary models, indicating a favorable scaling trend.

On Earth-Bench, HiRS-Agent substantially outperforms the corresponding vanilla Qwen3 backbones under both AP and IF. For Qwen3-4B, it improves Accuracy from 15.73/10.08 to 43.95/45.56, Tool-Exact-Match from 0.00/8.63 to 31.67/34.64, and Param-Match from 0.00/4.21 to 19.54/20.81, while reducing Efficiency from 3.9747/ 5.4554 to 2.5059/3.3013. Similar gains are observed on Qwen3-8B, where HiRS-Agent reaches 48.39/53.62 Accuracy.

On ThinkGeo, Table~\ref{tab:main_table2} reports a compact subset of the most representative metrics for comparison, including three step-level metrics (Inst., Tool., and Arg.) and two end-to-end metrics (Ans. and Ans\_I); the full metric set is provided in the supplementary material. Under the official evaluation protocol and tool library, HiRS-Agent consistently improves the corresponding Qwen3 backbones even without task-specific optimization on ThinkGeo. For Qwen3-4B, Inst./Tool./Arg. improve from 18.35/8.54/1.24 to 73.73/47.87/8.51, while Ans./Ans\_I increase from 6.07/7.79 to 11.28/13.77. For Qwen3-8B, the corresponding metrics improve from 20.98/13.36/3.26 and 7.67/8.68 to 77.97/59.57/11.70 and 12.09/14.75, respectively. Notably, HiRS-Agent achieves the best Ans. score among all compared methods.

Overall, the results on both benchmarks show that HiRS-Agent provides strong system-level gains for long-horizon RS agents and remains effective across different tool environments.

\subsection{RS Domain Adaptation Result}
\label{sec:domain_adapt}

We evaluate whether Expert-tuning improves RS expertise while preserving general capability. Table~\ref{tab:rs_expertise_and_general} reports results on RS-EXPERT-BENCHMARK and three general benchmarks.

Expert-tuning yields clear gains over both the Base model and the Prompt variant on RS-EXPERT-BENCHMARK. For Qwen3-4B, the overall score improves from 73.35 to 87.60, with consistent gains on both the in-domain and out-of-domain splits. This indicates that Expert-tuning improves RS knowledge and workflow understanding rather than merely benefiting from prompt reformulation.

At the same time, general capability is largely preserved. MMLU-Redux remains nearly unchanged, while MATH-500 and Multi-IF show only moderate drops. Overall, Expert-tuning achieves a favorable specialization--retention trade-off, substantially improving RS expertise without severely sacrificing general reasoning ability. Results on Qwen3-8B are provided in the supplementary material.

\subsection{Ablation Studies}
\label{sec:ablation}

\textbf{Training Ablation.}
We evaluate each training stage under the fixed HiRS-Agent
architecture. As shown in Table~\ref{tab:ablation_earth}(a),
Expert-tuning improves Exact from 11.47/12.85 to 18.07/19.78 and
Accuracy from 35.48/33.06 to 40.73/39.52, indicating that supervised
workflow alignment improves both exact tool execution and final-task
correctness. VG-HRL further improves all reported metrics, achieving
41.10/46.34 In-Order, 31.67/34.64 Exact, and 43.95/45.56 Accuracy.
These results show that hierarchical policy optimization complements
Expert-tuning by further improving ordered and exact execution.

\textbf{Architecture Ablation.}
We isolate the architectural components using the same trained
checkpoint and toolset. As shown in Table~\ref{tab:ablation_earth}(b),
adding verification to the flat agent improves Exact from 6.84/19.42
to 29.05/30.14, demonstrating its effectiveness in correcting
intermediate execution errors. The generic hierarchy with verification
improves In-Order to 32.47/34.92, while further introducing
RS-specialist grouping raises Accuracy from 30.24/33.47 to
43.95/45.56. Conversely, removing verification from the full system
reduces Exact by 11.30/12.43 and Accuracy by 4.03/2.82, confirming
that hierarchical specialization and step-level verification provide
distinct and complementary benefits.

\begin{table}[t]
	\centering
	\small
	\caption{
	Core ablation results on Earth-Bench using Qwen3-4B.
	Complete metrics are provided in the supplementary material.
	}
	\label{tab:ablation_earth}
	\begin{tabular}{l*{6}{c}}
		\toprule
		\multirow{2}{*}{Config.} &
		\multicolumn{2}{c}{In-Order$\uparrow$} &
		\multicolumn{2}{c}{Exact$\uparrow$} &
		\multicolumn{2}{c}{Accuracy$\uparrow$} \\
		\cmidrule(lr){2-3}
		\cmidrule(lr){4-5}
		\cmidrule(lr){6-7}
		& AP & IF & AP & IF & AP & IF \\
		\midrule

		\multicolumn{7}{l}{\itshape(a) Training stages} \\

		Base &
		\underline{22.31} & 19.43 &
		11.47 & 12.85 &
		35.48 & 33.06 \\

		Base + ET &
		19.73 & \underline{23.68} &
		\underline{18.07} & \underline{19.78} &
		\underline{40.73} & \underline{39.52} \\

		\rowcolor{gray!15}
		Base + ET + VG &
		\textbf{41.10} & \textbf{46.34} &
		\textbf{31.67} & \textbf{34.64} &
		\textbf{43.95} & \textbf{45.56} \\

		\midrule
		\multicolumn{7}{l}{\itshape(b) Architecture components} \\

		Flat &
		16.65 & 18.92 &
		6.84 & 19.42 &
		16.53 & 20.16 \\

		Flat + V &
		28.77 & 30.71 &
		\underline{29.05} & \underline{30.14} &
		35.89 & 31.45 \\

		Generic + V &
		\underline{32.47} & \underline{34.92} &
		23.82 & 23.32 &
		30.24 & 33.47 \\

		RS-Hier. w/o V &
		21.13 & 23.23 &
		20.37 & 22.21 &
		\underline{39.92} & \underline{42.74} \\

		\rowcolor{gray!15}
		Full HiRS-Agent &
		\textbf{41.10} & \textbf{46.34} &
		\textbf{31.67} & \textbf{34.64} &
		\textbf{43.95} & \textbf{45.56} \\

		\bottomrule
	\end{tabular}

		\par
	\scriptsize
	\textit{Note.} ET, VG, V, and RS-Hier.\ denote Expert-tuning,
	VG-HRL, the verifier, and the RS-specialist hierarchy, respectively;
	bold and underline mark the best and second-best results per panel.
\end{table}

\section{Conclusion}
In this paper, we present HiRS-Agent, a hierarchical multi-agent framework for long-horizon RS task solving. By organizing the system into a Manager Layer and a Specialist Layer, HiRS-Agent enables structured task decomposition, dynamic coordination, and domain-specialized tool execution aligned with the multi-stage, interdependent nature of RS workflows. To improve reliability, HiRS-Agent aligns RS expertise with workflows and optimizes policies through step-level verification, expert-tuning, and verification guided HRL. Experiments on Earth-Bench and ThinkGeo show that HiRS-Agent consistently improves long-horizon tool-use capability and final-task performance on lightweight open-source backbones. Notably, even when built on a lightweight 4B open-source backbone, HiRS-Agent is able to outperform the strongest closed-source models in certain settings.
These results highlight that, rather than relying on model scale alone, explicit modeling of RS workflow structure and stage dependency is critical for achieving reliable long-horizon execution. Future work will extend HiRS-Agent toward broader tools and more realistic scenarios.

\begin{acks}
This work has been funded by the National Natural Science Foundation of China under Grant 62301063.
\end{acks}

\bibliographystyle{ACM-Reference-Format}
\bibliography{sample-base}

\appendix

\end{document}